# An improved approach to attribute reduction with covering rough sets

Changzhong Wang [a,*], Baiqing Sun [b], Qinhua Hu [c]

[a]Department of Mathematics, Bohai University, Jinzhou, 121000, P.R.China

[b]Economy and management school, Harbin institute of technology,
Harbin, 150001, P.R.China

[c] Power Engineering, Harbin Institute of Technology, 150001 Harbin, PR China

**Abstract:** Attribute reduction is viewed as an important preprocessing step for pattern recognition and data mining. Most of researches are focused on attribute reduction by using rough sets. Recently, Tsang et al. discussed attribute reduction with covering rough sets in the paper [E. C.C. Tsang, D. Chen, Daniel S. Yeung, Approximations and reducts with covering generalized rough sets, Computers and Mathematics with Applications 56 (2008) 279–289], where an approach based on discernibility matrix was presented to compute all attribute reducts. In this paper, we provide an improved approach by constructing simpler discernibility matrix with covering rough sets, and then proceed to improve some characterizations of attribute reduction provided by Tsang et al. It is proved that the improved discernible matrix is equivalent to the old one, but the computational complexity of discernible matrix is greatly reduced.

## 1. Introduction

Attribute reduction and feature selection have become one of the important steps for pattern recognition and machine learning tasks. Classical rough set theory [19] is a mathematical tool for handling data sets with imprecision and uncertainty. It can be employed to study attribute reduction and feature selection in information systems. Equivalence relations are the mathematical basis for the rough set theory. Based on equivalence relations, objects of a universe can be partitioned into exclusive equivalence classes, which form the basic information granules to approximate arbitrary subset of the universe. The main idea of rough sets is to remove redundant information in data and to make correct decision or classification. Rough set theory has attracted wide attention on the research areas in both of the theory and its applications.

---

* Corresponding author. Tel.: +86 416 3400145; fax: +86 416 3400149.
Email address: changzhongwang@126.com (C. Wang).

Many types of attribute reductions have been proposed based on classical rough sets such as possible reduct, approximate reduct, α-reduct, μ-decision reduct and so on [9, 17,24]. Kryszkiewicz [9] reviewed and compared these five types of attribute reducts in inconsistent systems. In fact, only two of them, possible reduct and μ-decision reduct, are essential because the others are just equivalent to one of them, respectively. In addition, some other reduction methods based on classical rough sets were also proposed in [13,14,22,23,31].

However, equivalence relation in classical rough set theory is still restrictive for many applications, as it is only suitable for handling discrete variables and cannot directly deal with continuous or real-value data. There are large amount of continuous data in real-life applications. For example, a lot of numerical data are faced with on performance analysis and equipment condition monitoring and diagnosis in power systems [20]. When dealing with such numerical attributes by using classical rough sets, they are often discretized firstly into symbol-type attributes as a pretreatment [18]. This type of conversion will bring a major drawback of information loss, thus affecting the accuracy of extracted rules [8]. In order to solve this problem, scholars have proposed a series of extensions of the rough set model [1-8,10-12,15,16,26-30,32-35]. On the basis of the concept of cover on a universe, Pomykala et al [21] introduced the concepts of lower and upper covering approximation operators in the universe. Afterward many authors conducted detailed study of properties of covering approximation operators [1,15,21,28,32-35]. However, few people employ covering rough sets to make research on attribute reduction. Zhu [33] et al investigated reduction of covering elements based on covering rough sets. The reduction of covering elements is referred as to a means to get rid of excessive covering elements in a cover under the condition that keeps the upper and lower approximation of an arbitrary subset invariant. So what they meant about reduction is not the strict sense of attribute reduction. In [28] a pioneering work related to attribute reduction with covering rough sets was conducted, where the authors constructed discernible matrix and analyzed its some important properties. In view of the discernible matrix, an approach to compute all the reducts was developed. However, the formula for computing discernible matrix is very complicated, and so should be difficultly applied in practice.

In this paper, we revisit the discernible-matrix approach to attribute reduction with covering rough sets. More concretely, we reconstruct discernible matrix of attribute reduction. Compared with the approach in [28], the computational complexity of improved approach is lower. In addition, we improve the theorems in [28] that describe the properties of discernible matrix and attribute reduction with covering rough sets. The theory improved here is helpful in establishing a basic foundation of covering rough sets and broadening its applications.

The remainder of this paper is structured as follows. In Section 2, we recall and define some basic notions related to covering rough sets. In Section 3, we reconstruct discernible matrix of attribute reduction based on covering rough sets and improve some characterizations of the structure and properties of attribute reduction. In Section 4, we conclude the paper with a summary.

## 2. Some notions related to covering rough sets

Attribute reduction is an important application field of rough set theory. However, in real world there are lots of data sets that cannot be handled well by classical rough sets. In light of this, similarity relation rough sets [26], dominance rough sets [3,4], and even neighborhood rough sets [6,10] were developed one by one. All these models induce covers of a universe, instead of partitions, and thus can be categorized into covering rough sets, which are more general than classical rough sets and can handle more complex tasks.

Granulating information in data sets is the basis of rough set theory. The granulated information forms elementary information granules to approximately describe arbitrary concepts in approximation spaces. Covering rough set theory employs the notion of covers to granulate information in data sets.

**Definition 2.1**[1] Let $U$ be a nonempty and finite set of objects, where $U = \{x_1, x_2, ..., x_n\}$ is called a universe of discourse. $C = \{K_1, K_2, ..., K_m\}$ is a family of nonempty subsets of $U$, and $\bigcup_{i=1}^{m} K_i = U$. We say $C$ is a cover of $U$, $K_i$ is a covering element, and the ordered pair $(U, C)$ is a covering approximation space.

**Definition 2.2** [28] Suppose $U$ is a finite universe and $C=\{K_1, K_2, ..., K_m\}$ is a cover of $U$. For any $x \in U$, let $C(x) = \bigcap\{K_j \in C : x \in K_j\}$, then $Cov(C) = \{C(x) : x \in U\}$ is also a cover of $U$, we call it the induced cover of $C$.

$C(x)$ is the minimal descriptive subset containing $x$. This means $C(x)$ cannot be written as the union of other elements in $Cov(C)$. Thus $C(x)$ can be seen as the information granule of $x$ with respect to $C$ and $Cov(C)$ can be viewed as a set of information granules on $U$. These information granules are minimal covering elements associated with objects. $Cov(C) = C$ if and only if $C$ is a partition. For any $x, y \in U$, if $y \in C(x)$, then $C(y) \subseteq C(x)$. So if $y \in C(x)$ and $x \in C(y)$, then $C(x) = C(y)$. The relationships between information granules have the following properties.

(1) Reflexivity: $\forall x \in U$, $x \in C(x)$.

(2) Anti-symmetry: if $y \in C(x)$ and $x \in C(x)$, then $C(x) = C(y)$.

(3) Transitivity: $\forall x, y, z \in U$, if $x \in C(y)$ and $y \in C(z)$, then $x \in C(z)$.

In classification and regression learning, we are usually confronted with the task of approximating some concepts with provided knowledge. With information granules in covering approximation spaces, any concepts can be approximated.

**Definition 2.3** [28] Let $(U, C)$ be a covering approximation space. $X \subseteq U$ is an arbitrary subset of the universe. The covering lower and upper approximations of X are defined as $\underline{C}(X) = \bigcup\{C(x) : C(x) \subseteq X\}$, $\overline{C}(X) = \bigcup\{C(x) : C(x) \cap X \neq \emptyset\}$.

**Definition 2.4** [28] Suppose $U$ is a finite universe and $\Delta = \{C_i : i = 1, ..., m\}$ is a family of covers of $U$. For any $x \in U$, let $\Delta(x) = \bigcap\{C_i(x) \in Cov(C_i) : i = 1, 2 \cdots, m\}$, then $Cov(\Delta) = \{\Delta(x) : x \in U\}$ is also a cover of $U$, we call it the induced cover of $\Delta$.

Clearly, $\Delta(x)$ is the intersection of all the covering elements including $x$ in all covers, and so the minimal descriptive set containing $x$ in $Cov(\Delta)$. Similarly, $\Delta(x)$ can be viewed as the information granule of $x$ with respect to $\Delta$ and $Cov(\Delta)$ can be viewed as a set of information granules with respect to $\Delta$. If every cover in $\Delta$ is a partition, then $Cov(\Delta)$ is also a partition and $\Delta(x)$ is the equivalence class containing $x$. Each information granule in $Cov(\Delta)$ cannot be written as the union of other granules. For any $x, y \in U$, if $y \in \Delta(x)$, then $\Delta(x) \supseteq \Delta(y)$. So if $y \in \Delta(x)$ and $x \in \Delta(y)$, then $\Delta(x) = \Delta(y)$. The relationships between information granules in $Cov(\Delta)$ also have such properties as reflexivity, anti-symmetry and transitivity.

**Definition 2.5** Let $U$ be a universe and $\Delta = \{C_i : i = 1, ..., n\}$ a family of covers on $U$. Then $(U, \Delta)$ is called a covering information system; $\Delta$ is called a conditional covering (attribute) set.

**Definition 2.6** Let $(U, \Delta)$ be a covering information system and $C_i \in \Delta$. $C_i$ is called superfluous in $\Delta$ if $Cov(\Delta - \{C_i\}) = Cov(\Delta)$, i.e. $(\Delta - \{C_i\})(x) = \Delta(x)$ for any $x \in U$. Otherwise, $C_i$ is called indispensable in $\Delta$. For any subset $\mathbf{P} \subseteq \Delta$, $\mathbf{P}$ is called a reduct of $\Delta$ if each element in $\mathbf{P}$ is indispensable in $\mathbf{P}$ and $Cov(\mathbf{P}) = Cov(\Delta)$. The collection of all indispensable elements in $\Delta$ is called the core of $\Delta$, denoted as $Core(\Delta)$.

Definitions 2.5 and 2.6 are natural extensions of the corresponding concepts in classical rough set theory by substituting equivalence relations with covers. It can be seen from the two definitions that the purpose for reducing conditional covering set is to find a minimal covering subset that keeps original information granularity invariant.

## 3. Attribute reduction based on discernbility matrix

In this section, we first develop some theorems to describe discernbility between objects. Then, we reconstruct the discernible matrix of attribute reduction based on covering rough sets and improve some characterizations of basic properties of attribute reduction.

Let $U = \{x_1, x_2, ..., x_n\}$ be a universe, $\Delta = \{C_1, C_2, \cdots C_n\}$ be a family of covers on $U$, $C_k \in \Delta$. For any $x_i, x_j \in U$, if $x_j \notin C_k(x_i)$, then we say $x_i$ and $x_j$ can be distinguished by $C_k$. This statement accords to the corresponding views in classical rough sets.

**Proposition 3.1.** $Cov(\mathbf{P}) = Cov(\Delta) \Leftrightarrow \Delta(x) = \mathbf{P}(x)$, $\forall x \in U$. □

The proposition presents an equivalence condition to judge whether two covers are equal and shows the fact that two covers are equal if and only if their induced granularities are equal.

**Theorem 3.2.** Let $\Delta = \{C_1, C_2, \cdots C_n\}$ be a family of covers on $U$, $C_i \in \Delta$, $x \in U$. Then $C_i$ is an indispensable cover if and only if there exists $y \in U$ such that $y \notin \Delta(x) \Rightarrow y \in \{\Delta - \{C_i\}\}(x)$.

**Proof.** Straightforward. □

The above theorem implies that an indispensable cover can be characterized by the discernibility between objects. That is to say, $C_i$ is an indispensable cover if and only if $y$ not belonging to the neighborhood of $x$ with respect to $\Delta$ implies $y$ belonging to the neighborhood of $x$ with respect to $\Delta - \{C_i\}$. This implies that $C_i$ is a sole cover that can distinguish the two objects.

**Theorem 3.3.** Let $\Delta = \{C_1, C_2, \cdots C_n\}$ be a family of covers on $U$. For $\forall x, y \in U$, $y \notin \Delta(x)$ if and only if there is at least a cover $C_i \in \Delta$ such that $y \notin C_i(x)$.

**Proof.** Straightforward. □

This theorem shows that if two objects can be distinguished under the granularity level of $\Delta(x)$, then there is at least a cover $C_i \in \Delta$ such that the two objects are also distinguished under the granularity level of $C_i(x)$, and vice versa. This means that if two objects are not in the same original information granule, i.e. one object is not in the neighborhood of another object with respect to $\Delta$, we can find at least an attribute (cover) to distinguish them.

**Theorem 3.4.** (Judgment theorem of attribute reduction). Let $\Delta = \{C_1, C_2, \cdots C_n\}$ be a family of covers on $U$, $\mathbf{P} \subseteq \Delta$. Then $Cov(\mathbf{P}) = Cov(\Delta)$ if and only if for $\forall x, y \in U$, if $y \notin \Delta(x)$, then $y \notin \mathbf{P}(x)$.

**Proof.** If $Cov(\mathbf{P}) = Cov(\Delta)$, then by Proposition 3.1 we have $\Delta(x) = \mathbf{P}(x)$ for every $x \in U$. Hence if $y \notin \Delta(x)$, then $y \notin \mathbf{P}(x)$. On the other hand, since $\Delta(x) \subseteq \mathbf{P}(x)$, it follows that $y \notin \mathbf{P}(x) \Rightarrow y \notin \Delta(x)$ is always true. So $y \notin \Delta(x) \Rightarrow y \notin \mathbf{P}(x)$ is equivalent to $y \notin \Delta(x) \Leftrightarrow y \notin \mathbf{P}(x)$. By Proposition 3.1, we have $Cov(\mathbf{P}) = Cov(\Delta)$. □

As mentioned above, the objective of attribute reduction is to find out minimal subsets of conditional covering set $\Delta$ that keeps invariant the minimal descriptive set of every object with respect to $\Delta$. This theorem shows attribute reduction must keep invariant the original discernbility between any two objects. If two objects are distinguished under the original granularity level, they have to be distinguished under lower level of granularity induced by the candidate-attribute set of reducts. In order to search for all the reducts, we define discernibility matrix according to Theorems 3.3 and 3.4 as follows.

**Definition 3.1.** Let $(U,\Delta)$ be a covering information system. Suppose $U = \{x_1, x_2 \ldots, x_n\}$, we denote by $M(U,\Delta)$ a $n \times n$ matrix $(c_{ij})$, called the discernibility matrix of $(U,\Delta)$ and defined as $c_{ij} = \{C \in \Delta : x_j \notin C(x_i)\}$ for $x_i, x_j \in U$. Clearly, we have $c_{ii} = \varnothing$ for any $x_i \in U$.

As an attribute and its induced cover are uniquely determined by each other, the discernibility matrix gives the description of all the attribute subsets that can distinguish any two objects. If two objects don't belong to one information granule at original level of granularity, then they must be distinguished by some attributes. This idea is consistent with the viewpoint in classical rough sets. Besides, covering and classical rough sets bear some formal resemblance between discernibility matrices. Thus the proposed approach is a generalization of classical rough sets.

In [28] the reduction method based on discernibility matrix was also proposed to compute all the reducts. For the sake of comparison, let us review the definition of discernibility matrix introduced there.

**Definition 3.2** [28]. Let $U = \{x_1, x_2, \ldots, x_n\}$. By $M(U,\Delta)$ we denote a $n \times n$ matrix $(w_{ij})$, called the discernibility matrix of $(U,\Delta)$ such that for $x_i, x_j \in U$,

$$w_{ij} = \begin{cases} \varnothing, & \Delta(x_i) = \Delta(x_j) \\ \{C \in \Delta \mid C(x_i) \subset C(x_j)\}, & \Delta(x_i) \subset \Delta(x_j) \text{ or } \{C \in \Delta \mid C(x_j) \subset C(x_i)\}, \Delta(x_j) \subset \Delta(x_i) \\ \{C \in \Delta \mid C(x_i) \not\subset C(x_j) \wedge C(x_j) \not\subset C(x_i)\} \cup \{C_s \wedge C_t \mid C_s(x_i) \subset C_s(x_j) \wedge C_t(x_j) \subset C_t(x_i)\}, \Delta(x_i) \not\subset \Delta(x_j) \wedge \Delta(x_j) \not\subset \Delta(x_i) \end{cases}$$

In fact, the two types of discernbility matrices are equivalent. Next, we present the proof of their equivalence.

**Theorem 3.5.** The discernbility matrices in Definition 3.1 and 3.2 are equivalent.

**Proof.** To prove the statement is true, we only need to verify $c_{ij} = w_{ij}$ for any $x_i, x_j \in U$.

(1) If $\Delta(x_i) = \Delta(x_j)$, then $x_j \in \Delta(x_i)$ and $x_i \in \Delta(x_j)$ by the reflexivity of $\Delta(x_i)$ and $\Delta(x_j)$. According to Definition 3.1, we have $c_{ij} = \varnothing$, which implies that $c_{ij} = w_{ij}$.

(2) If $\Delta(x_i) \subset \Delta(x_j)$, it follows from the reflexivity and transitivity of information granules that $x_j \notin \Delta(x_i)$ and $x_i \in \Delta(x_j)$. This implies that $\Delta(x_i) \subseteq \Delta(x_j)$ for any $C_i \in \Delta$. Let $C \in c_{ij}$, it follows from Definition 3.1 that $x_j \notin C(x_i)$. This, together with $\Delta(x_i) \subseteq \Delta(x_j)$, means that $C(x_i) \subset C(x_j)$. From Definition 3.2 we get $C \in w_{ij}$, which implies $c_{ij} \subseteq w_{ij}$. On the other hand, let $C \in w_{ij}$, by Definition 3.2 we know $C(x_i) \subset C(x_j)$. It follows from the reflexivity of information granules that $x_j \notin C(x_i)$. This by Definition 3.1 means that $C \in c_{ij}$. Thus $c_{ij} \supseteq w_{ij}$, as desired.

(3) If $\Delta(x_i) \not\subset \Delta(x_j) \wedge \Delta(x_j) \not\subset \Delta(x_i)$, let $C_s \in c_{ij}$. By Definition 3.1 we have $x_j \notin C_s(x_i)$, and so $C_s(x_j) \not\subset C_s(x_i)$ by the reflexivity. This means $\Delta(x_j) \not\subset \Delta(x_i)$. Considering another fact $\Delta(x_i) \not\subset \Delta(x_j)$, there are two possible cases for the cover $C_s$.

(i) $C_s(x_i) \not\subset C_s(x_j)$; (ii) $C_s(x_i) \subseteq C_s(x_j)$.

From the first case and the fact $C_s(x_j) \not\subset C_s(x_i)$, we can get $C_s(x_i) \not\subset C_s(x_j) \wedge C_s(x_j) \not\subset C_s(x_i)$. It follows from Definition 3.2 that $C_s \in w_{ij}$. If $C_s$ meets the second case, we have $C_s(x_i) \subseteq C_s(x_j) \wedge C_s(x_j) \not\subset C_s(x_i)$. By the fact $\Delta(x_i) \not\subset \Delta(x_j)$, there must exist another cover $C_t \in \Delta$ such that $C_t(x_i) \not\subset C_t(x_j)$. Thus we get $C_t(x_i) \not\subset C_t(x_j) \wedge C_s(x_j) \not\subset C_s(x_i)$. This implies by Definition 3.2 that $C_s \wedge C_t \in w_{ij}$. So $c_{ij} \subseteq w_{ij}$. On the other hand, let $C \in w_{ij}$, then $C$ meets $C(x_i) \not\subset C(x_j) \wedge C(x_j) \not\subset C(x_i)$. This means $x_j \notin C(x_i)$ by the transitivity. It follows from Definition 3.1 that $C \in c_{ij}$. Let $C_s \wedge C_t \in w_{ij}$, then by Definition 3.2 we can get t $C_t(x_i) \not\subset C_t(x_j) \wedge C_s(x_j) \not\subset C_s(x_i)$, which implies that $x_j \notin C_s(x_i)$ and $x_i \notin C_s(x_j)$. This by Definition 3.1 means that $C_s, C_t \in c_{ij}$. So $c_{ij} \supseteq w_{ij}$, as desired. □

The above theorem shows the two types of discernible matrices are equivalent to each other. This means that we can get the same results if they are employed to compute reducts of identical data sets. However, it can be easily observed that the proposed formula for computing discernible matrix in Definition 3.1 is simpler than that introduced in Definition 3.2. Now, let us analyze the computational complexity of them. As the complexity of matrix is $o(n^2)$, the time complexity of the discernibility matrix in Definition 3.2 is $o(m^2 \cdot n^2)$, while the time complexity in Definition 3.1 is $o(m \cdot n^2)$, where $n$ and $m$ are the numbers of samples and attributes, respectively. This justifies that our proposed method is simpler than the method in [28]. From theoretical viewpoint, Our study on this topic plays the same important role as the researches [25,30] in traditional and generalized rough sets.

The following theorem is used to study the properties of the discernibility matrix.

**Theorem 3.6.** Let $M(U, \Delta) = (c_{ij})$ be the discernibility matrix of $(U, \Delta)$ and $\Delta = \{C_1, C_2, \cdots C_n\}$. Then the following statements hold:

(1) $C_l \in c_{ij} \Leftrightarrow x_j \notin C_l(x_i)$;

(2) $c_{ii} = \emptyset$; $i \leq 1, 2, \ldots n$;

(3) $c_{ij} \subseteq c_{it} \cup c_{tj}$, $(i \neq j, i, j, t \leq 1, 2 \ldots n)$.

**Proof.** (1) Straightforward.

(2) For $\forall x_i \in U$, by the definition of $C(x)$ we have $x_i \in C_l(x_i)$ for any $C_l \in \Delta$. It follows from Definition 3.1 that $C_l \notin c_{ii}$. Thus $c_{ii} = \emptyset$.

(3) Let $C \in c_{ij}$, then $x_j \notin C(x_i)$. Suppose that $x_t \in C(x_i)$ and $x_j \in C(x_t)$, then by the definition of $C(x)$ we have $x_j \in C(x_i)$. This is equivalent to that $x_j \notin C(x_i) \Rightarrow x_t \notin C(x_i)$ or $x_j \notin C(x_t)$, which implies that $C \in c_{ij} \Rightarrow C \in c_{it}$ or $C \in c_{tj}$. So $c_{ij} \subseteq c_{it} \cup c_{tj}$. □

This theorem illustrates that the properties of discernibility matrix are determined by the properties of covers. If $\Delta$ is a family of equivalence relations, $M(U,\Delta)$ is the discernibility matrix of the corresponding information system in Pawlak's rough set theory.

In [28], Proposition 4.5 says that the core of attribute reduction is computed by

$$Core(\Delta) = \{C \in \Delta : c_{ij} = \{C\} \vee c_{ij} = \{C \wedge C_t\}, t = 1, 2, \cdots, m; i, j \leq n\}.$$

Here we improve it based on the discernible matrix in Definition 3.1 as follows.

**Theorem 3.7.** $Core(\Delta) = \{C \in \Delta : c_{ij} = \{C\}, i, j \leq n\}$.

**Proof.** Suppose $C \in Core(\Delta)$, then $Cov(\Delta) \neq Cov(\Delta - \{C\})$. By Proposition 3.1 and Theorem 3.2, it follows that there exist $x_i, x_j \in U$ such that $x_j \notin \Delta(x_i)$, but $x_j \in \{\Delta - \{C\}\}(x_i)$. Obviously, there is only $C \in \Delta$ satisfying $x_j \notin C(x_i)$. By Definition 3.1, $c_{ij} = \{C\}$. Hence $Core(\Delta) \subseteq \{C \in \Delta : c_{ij} = \{C\}, i, j \leq n\}$. Conversely, if $c_{ij} = \{C\}$ for $x_i, x_j \in U$, then $C \in Core(\Delta)$ by Theorem 3.2. Hence $Core(\Delta) \supseteq \{C \in \Delta : c_{ij} = \{C\}, i, j \leq n\}$. Therefore $Core(\Delta) = \{C \in \Delta : c_{ij} = \{C\}, i, j \leq n\}$. □

The following theorem is an improved version of Proposition 4.6 in [28].

**Theorem 3.8.** Let $\mathbf{P} \subseteq \Delta$, then $Cov(\mathbf{P}) = Cov(\Delta)$ if and only if $\mathbf{P} \cap c_{ij} \neq \emptyset$ for any $c_{ij} \neq \emptyset$.

**Proof.** $\Rightarrow$ Assume that $\exists i_0, j_0 \leq n$, $c_{i_0 j_0} \neq \emptyset$, but $\mathbf{P} \cap c_{i_0 j_0} = \emptyset$. Therefore, for $\forall C \in \mathbf{P}$ we have $x_{j_0} \in C(x_{i_0})$, which implies $x_{j_0} \in \mathbf{P}(x_{i_0})$. Since $Cov(\mathbf{P}) = Cov(\Delta)$, by Proposition 3.1 we have $x_{j_0} \in \Delta(x_{i_0})$. Thus $x_{j_0} \in C(x_{i_0})$ for any $C \in \Delta$, which implies $c_{i_0 j_0} = \emptyset$. This is a contradiction to the assumption. So $\mathbf{P} \cap c_{ij} \neq \emptyset$ for any $c_{ij} \neq \emptyset$.

$\Leftarrow$ If $\mathbf{P} \cap c_{ij} \neq \emptyset$ for $x_i, x_j \in U$ satisfying $c_{ij} \neq \emptyset$, we suppose $C \in \mathbf{P} \cap c_{ij}$, then we have $x_j \notin C(x_i) \Rightarrow x_j \notin \Delta(x_i) \Rightarrow x_j \notin \mathbf{P}(x_i)$. By Theorem 3.4 we have $Cov(\mathbf{P}) = Cov(\Delta)$. □

Let $\Delta = \{C_1, C_2, \cdots C_n\}$ be a family of covers on $U$. $f(U, \Delta)$ is a function on $(U, \Delta)$ and defined as $f(U, \Delta) = \wedge\{\vee(c_{ij})\}, (i, j \leq n, c_{ij} \neq \emptyset)$, where $\vee(c_{ij})$ represents the disjunction operation among elements in $c_{ij}$. By using the function, we can compute all the reducts of covering information systems. The following example is employed to compare our idea with that in [28].

**Example 3.1.** Let consider an example of *house evaluation problem* provided in [28]. For a detailed introduction to the example, the reader can refer to the reference. Suppose $U = \{x_1, x_2..., x_9\}$ is a set of nine houses, E = {price; color; structure; surrounding} is a set of attributes. For each of the four attributes, we can get a cover of $U$, denoted by $C_1$, $C_2$, $C_3$, $C_4$, respectively. The four covers are listed as follows.

$$C_1 = \{\{x_1,x_2,x_4,x_5,x_7,x_8\},\{x_2,x_5,x_8\},\{x_2,x_3,x_5,x_6,x_8,x_9\}\}.$$

$$C_2 = \{\{x_1,x_2,x_3,x_4,x_5,x_6\},\{x_4,x_5,x_6,x_7,x_8,x_9\}\}.$$

$$C_3 = \{\{x_1,x_2,x_3\},\{x_4,x_5,x_6,x_7,x_8,x_9\},\{x_7,x_8,x_9\}\}.$$

$$C_4 = \{\{x_1,x_2,x_4,x_5\},\{x_2,x_3,x_5,x_6\},\{x_4,x_5,x_7,x_8\},\{x_5,x_6,x_8,x_9\}\}.$$

Let $\Delta = \{C_1, C_2, C_3, C_4\}$, then

$C_1(x_1) = C_1(x_4) = C_1(x_7) = \{x_1,x_2,x_4,x_5,x_7,x_8\}, C_1(x_2) = C_1(x_5) = C_1(x_8) = \{x_2,x_5,x_8\},$

$C_1(x_3) = C_1(x_6) = C_1(x_9) = \{x_2,x_3,x_5,x_6,x_8,x_9\}.$

$C_2(x_1) = C_2(x_2) = C_2(x_3) = \{x_1,x_2,x_3,x_4,x_5,x_6\}, C_2(x_4) = C_2(x_5) = C_2(x_6) = \{x_4,x_5,x_6\},$

$C_2(x_7) = C_2(x_8) = C_2(x_9) = \{x_4,x_5,x_6,x_7,x_8,x_9\}.$

$C_3(x_1) = C_3(x_2) = C_3(x_7) = \{x_1,x_2,x_3\}, C_3(x_4) = C_3(x_5) = C_3(x_6) = \{x_4,x_5,x_6,x_7,x_8,x_9\},$

$C_3(x_7) = C_3(x_8) = C_3(x_9) = \{x_7,x_8,x_9\}.$

$C_4(x_1) = \{x_1,x_2,x_4,x_5\}, C_4(x_2) = \{x_2,x_5\}, C_4(x_3) = \{x_2,x_3,x_5,x_6\}, C_4(x_4) = \{x_4,x_5\}, C_4(x_5) = \{x_5\},$

$C_4(x_6) = \{x_5,x_6\}, C_4(x_7) = \{x_4,x_5,x_7,x_8\}, C_4(x_8) = \{x_5,x_8\}, C_4(x_9) = \{x_5,x_6,x_8,x_9\}.$

The discernibility matrix of $(U, \Delta)$ is as follows.

$$\begin{pmatrix} \varnothing & \varnothing & \{C_1,C_4\} & \{C_3\} & \{C_3\} & \{C_1,C_3,C_4\} & \{C_1,C_2,C_3,C_4\} & \{C_2,C_3,C_4\} & \{C_1,C_2,C_3,C_4\} \\ \{C_1,C_4\} & \varnothing & \{C_1,C_4\} & \{C_1,C_3,C_4\} & \{C_3\} & \{C_1,C_3,C_4\} & \{C_1,C_2,C_3,C_4\} & \{C_2,C_3,C_4\} & \{C_1,C_2,C_3,C_4\} \\ \{C_1,C_4\} & \varnothing & \varnothing & \{C_1,C_3,C_4\} & \{C_3\} & \{C_3\} & \{C_1,C_2,C_3,C_4\} & \{C_2,C_3,C_4\} & \{C_2,C_3,C_4\} \\ \{C_2,C_3,C_4\} & \{C_2,C_3,C_4\} & \{C_1,C_2,C_3,C_4\} & \varnothing & \varnothing & \{C_1,C_4\} & \{C_2,C_4\} & \{C_2,C_4\} & \{C_1,C_2,C_4\} \\ \{C_2,C_3,C_4\} & \{C_2,C_3,C_4\} & \{C_1,C_2,C_3,C_4\} & \{C_1,C_4\} & \varnothing & \{C_1,C_4\} & \{C_2,C_4\} & \{C_2,C_4\} & \{C_1,C_2,C_4\} \\ \{C_1,C_2,C_3,C_4\} & \{C_2,C_3,C_4\} & \{C_2,C_3,C_4\} & \{C_1,C_4\} & \varnothing & \varnothing & \{C_1,C_2,C_4\} & \{C_2,C_4\} & \{C_2,C_4\} \\ \{C_2,C_3\} & \{C_2,C_3\} & \{C_1,C_2,C_3\} & \{C_3\} & \{C_3\} & \{C_1,C_3\} & \varnothing & \varnothing & \{C_1,C_4\} \\ \{C_1,C_2,C_3,C_4\} & \{C_2,C_3,C_4\} & \{C_1,C_2,C_3,C_4\} & \{C_1,C_3,C_4\} & \{C_3\} & \{C_1,C_3,C_4\} & \{C_1,C_4\} & \varnothing & \{C_1,C_4\} \\ \{C_1,C_2,C_3,C_4\} & \{C_2,C_3,C_4\} & \{C_2,C_3,C_4\} & \{C_1,C_3,C_4\} & \{C_3\} & \{C_3\} & \{C_1,C_4\} & \varnothing & \varnothing \end{pmatrix}$$

and

$$f(U,\Delta) = \wedge\{\vee(c_{ij}): 1 \le j < i \le 9, c_{ij} \ne \varnothing\}$$

$$= (C_1 \vee C_4) \wedge C_3 \wedge (C_1 \vee C_3 \vee C_4) \wedge (C_1 \vee C_2 \vee C_3 \vee C_4) \wedge (C_2 \vee C_3 \vee C_4)$$
$$\wedge (C_2 \vee C_4) \wedge (C_1 \vee C_2 \vee C_4) \wedge (C_2 \vee C_3) \wedge (C_1 \vee C_2 \vee C_3) \wedge (C_1 \vee C_3)$$

$$= (C_1 \vee C_4) \wedge C_3 \wedge (C_2 \vee C_4)$$

$$= ((C_1 \wedge C_2) \vee C_4) \wedge C_3$$

$$= (C_1 \wedge C_2 \wedge C_3) \vee (C_4 \wedge C_3)$$

So $Red(\Delta) = \{\{C_3,C_4\},\{C_1,C_2,C_3\}\}$ and $Core(\Delta) = \{C_3\}$. □

The results are the same as ones in Ref.[28]. But the computational complexity is lower. So we can say that the proposed approch is simpler to compute all the reduts than the old one.

## 4. Conclusion

Covering rough sets are an important extension of classical rough sets. In this paper, we have redeveloped a relatively simple formula for computing discernible matrix. Although the proposed discernible matrix is equivalent to the one introduced in [28] and the results may be the same in computing attribute reducts, the computational complexity of our discernible matrix is lower. In addition, we have improved some characterizations of attribute reduction with covering rough sets. Compared with the results in [28], the improved ones are more concise, more profound to see through the nature of attribute reduction. These results obtained in the paper accord to the corresponding ones in classical rough sets and may help us develop more efficient approaches to attribute reductions, and so deal with more complex data sets.

**Acknowledgement**

This work was supported by the National Natural Science Foundation of China (No. 61070242), the Scientific Research Project of Department of Education of Hebei Province (No. 2009410), the Fundamental Research Funds for the Central Universities (No.HIT.NSRIF. 2010078).